\pdfoutput=1

\documentclass[11pt]{article}

\usepackage[]{ACL2023}

\usepackage{times}
\usepackage{latexsym}

\usepackage{graphicx}
\usepackage{tabularx}
\usepackage{soul}
\usepackage{booktabs}
\usepackage{caption}
\usepackage{subcaption}
\usepackage{tikz}
\usepackage{xcolor}
\definecolor{lightgray}{gray}{0.9}
\usetikzlibrary{arrows}
\usetikzlibrary{positioning}
\usetikzlibrary{shapes}
\usepackage{float}
\usepackage{hyperref}

\usepackage{titlesec}

\usepackage[T1]{fontenc}

\usepackage[utf8]{inputenc}

\usepackage{microtype}

\usepackage{inconsolata}

\usepackage{xcolor}
\usepackage{hyperref}
 \definecolor{darkblue}{rgb}{0, 0, 0.5}
  \hypersetup{colorlinks=true, citecolor=darkblue, linkcolor=darkblue, urlcolor=darkblue}

\usepackage{xstring}

\usepackage{color}

\usepackage[disable]{todonotes}
\newcommand\acb[1]{\textcolor{black}{#1}}

\definecolor{datablue}{RGB}{0,191,196}
\definecolor{gptgray}{RGB}{231,231,231}
\definecolor{filtergreen}{RGB}{153,216,201}

\title{Can LLMs Augment Low-Resource Reading Comprehension Datasets? Opportunities and Challenges}

 \author{\\ \textbf{Vinay Samuel}$^{\ddagger}$ $\quad$ \textbf{Houda Aynaou}$^{\S}$  $\quad$ \textbf{Arijit Ghosh Chowdhury}$^{\dag}$ $\quad$ \\ \textbf{Karthik Venkat Ramanan}$^{\dag}$ $\quad$ \textbf{Aman Chadha}$^{\clubsuit, \diamondsuit}$ \\ \\
 $^{\ddagger}$Carnegie Mellon University, $^\S$Georgia Institute of Technology, \\ $^{\dag}$University of Illinois Urbana-Champaign $^{\clubsuit}$Stanford University $^{\diamondsuit}$Amazon AI \vspace{0.2em}\\ 
\texttt{vsamuel@andrew.cmu.edu, haynaou3@gatech.edu, arijit10@gmail.com,}\\ \texttt{kv16@illinois.edu, hi@aman.ai}
 }
 
\begin{document}
\maketitle
\renewcommand{\thefootnote}{\fnsymbol{footnote}}
\footnotetext[1]{Work does not relate to position at Amazon.}
\renewcommand*{\thefootnote}{\arabic{footnote}}
\setcounter{footnote}{0}
\begin{abstract}
Large Language Models (LLMs) have demonstrated impressive zero shot  performance on a wide range of NLP tasks, demonstrating the ability to reason and apply commonsense. A relevant application is to use them for creating high quality synthetic datasets for downstream tasks. In this work, we probe whether GPT-4 can be used to augment existing extractive reading comprehension datasets. Automating data annotation processes has the potential to save large amounts of time, money and effort that goes into manually labelling datasets. In this paper, we evaluate the performance of GPT-4 as a replacement for human annotators for low resource reading comprehension tasks, by comparing performance after fine tuning, and the cost associated with annotation. This work serves to be the first analysis of LLMs as synthetic data augmenters for QA systems, highlighting the unique opportunities and challenges. Additionally, we release augmented versions of low resource datasets, that will allow the research community to create further benchmarks for evaluation of generated datasets.
\end{abstract}

\section{Introduction}
Machine reading comprehension (MRC) is a challenging NLP task where systems are designed to answer questions based on a given context. This task has significant practical value, as it answers user queries in diverse settings, from clinical contexts \cite{clinical_qa_1_bio_asq, clinincal_qa_2_pampari-etal-2018-emrqa, clinical_qa_3}, to customer support \cite{customer_support_1_-techqa} and policy interpretation \cite{policy_1_policy_qa}. BERT-based models \cite{bert_model_qa} have achieved state-of-the-art performance when trained with extensive data from datasets like SQuAD \cite{squad_2} and Natural Questions \cite{natural_questions}. However, their effectiveness diminishes in low-resource domains with limited datapoints \cite{low_resource_qa_bad}. This limitation becomes particularly pronounced in newly emerging fields such as COVID-19 \cite{covidqa_clinical_qa_4_moller-etal-2020}, where substantial annotated instances are often lacking.

Data augmentation has been instrumental in enhancing performance across numerous low-resource NLP tasks \cite{data_augmentation_NLP_survey, data_augmentation_2_nlu, data_augmentation_3_ner}. Yet, much of the work on data augmentation for QA \cite{synthetic_qa_1_roundtrip, synthetic_qa_2_unlabeled, synthetic_qa_3_robust, synthetic_qa_4_simple_semi_sup, synthetic_qa_5_simple_semi_sup}, hinges on the availability of unlabeled paragraphs from common sources, such as Wikipedia, to produce new context-question-answer instances. This approach poses a challenge for specialized and mission-critical domains where such unlabeled contexts are scarcely available. Bridging this gap, LLMs \cite{ gpt3} exhibit a capability to generate texts that closely resemble human-authored content \cite{gpt3, llm_like_human}. This potential of LLMs can be harnessed to generate both novel contexts and their corresponding question-answer pairs.

Addressing this gap, we introduce a GPT-4 \cite{openai2023gpt4} based data augmentation technique tailored for low-resource machine reading comprehension, specifically focusing on the extractive setting. Our approach begins by generating supplementary contexts, questions, and answers to augment training sets. To achieve this, we use in-context learning with passages, questions, and answers from the training set, ensuring minimal domain shift between the synthetically generated data and the original datasets

Subsequently, we adopt cycle-consistent filtering to isolate high-quality training instances. Empirical evaluations conducted on three pertinent real-world low-resource datasets CovidQA \cite{covidqa_clinical_qa_4_moller-etal-2020}, PolicyQA \cite{policy_1_policy_qa}, and TechQA \cite{customer_support_1_-techqa} reveal that our methodology improves the performance of BERT-based MRC on CovidQA by 23\% and on PolicyQA by 5\% in terms of exact match. Notably, our approach attains state-of-the-art results on CovidQA.

\section{Related Work}

Language models have played a key role in the creation of synthetic datasets for various NLP tasks. Models such as GPT-2 \cite{gpt2} and CTRL \cite{ctrl} have been applied to areas including general language understanding \cite{meng_data_aug_using_ctrl, gal}, classification \cite{data_aug_using_pretrained, to_the_rescue}, dialogue tasks \cite{gpt2_for_dialog_systems}, commonsense reasoning \cite{gpt2_for_commonsense}, and relation extraction \cite{gpt2_reln_extraction}, among others. Recently, large language models have significantly improved the quality and scope of synthetic dataset generation. They have been instrumental in augmenting datasets for tasks such as NLI and sentiment analysis \cite{core}, classification \cite{gpt3mix}, and even creating datasets for personalized dialogue generation \cite{personachatgen}, hate speech detection \cite{toxigen}, and textual similarity \cite{gpt3_embeddings} to name a few.

Most prior work in synthetic data generation for QA \cite{qa-data-aug-1, qa-data-aug-2, qa-data-aug-3, synthetic_qa_1_roundtrip} has concentrated on generating questions from Wikipedia passages to produce supplementary training examples. More recently, \citeauthor{quasi-swedish-gpt-3} introduced the use of GPT-3 for creating extra training data for Swedish multiple choice questions. Our approach is the first to utilize in-context learning with LLMs for synthesizing contexts, questions, and answers for low-resource MRC.
 
\section{Setup}

\subsection{Low Resource Datasets}

We utilize three reading comprehension datasets in our work: CovidQA, PolicyQA, and TechQA. These datasets cover diverse domains while having relatively small training sizes, making them well-suited for evaluating synthetic data augmentation techniques.

The CovidQA dataset \cite{covidqa_clinical_qa_4_moller-etal-2020} focuses on question answering related to the COVID-19 pandemic. It contains 2,019 question-answer pairs on topics such as virus transmission, public health interventions, and social impacts.

PolicyQA \cite{policy_1_policy_qa} contains 12,102 question-answer pairs about United States immigration and travel policies. The questions require reasoning about specific policy documents to determine the answer.

TechQA \cite{customer_support_1_-techqa} provides 1,808 examples related to technical support issues on computer networking, software, and hardware. The goal is to develop QA systems that can resolve technical problems automatically.

In summary, these three datasets cover the domains of healthcare, public policy, and technology, while having relatively small training set sizes between 1-10k examples. This makes them suitable testbeds for studying the effects of augmenting the training data through synthetic example generation.

\begin{figure*}[t]
 \centering
 \includegraphics[width=1\linewidth]{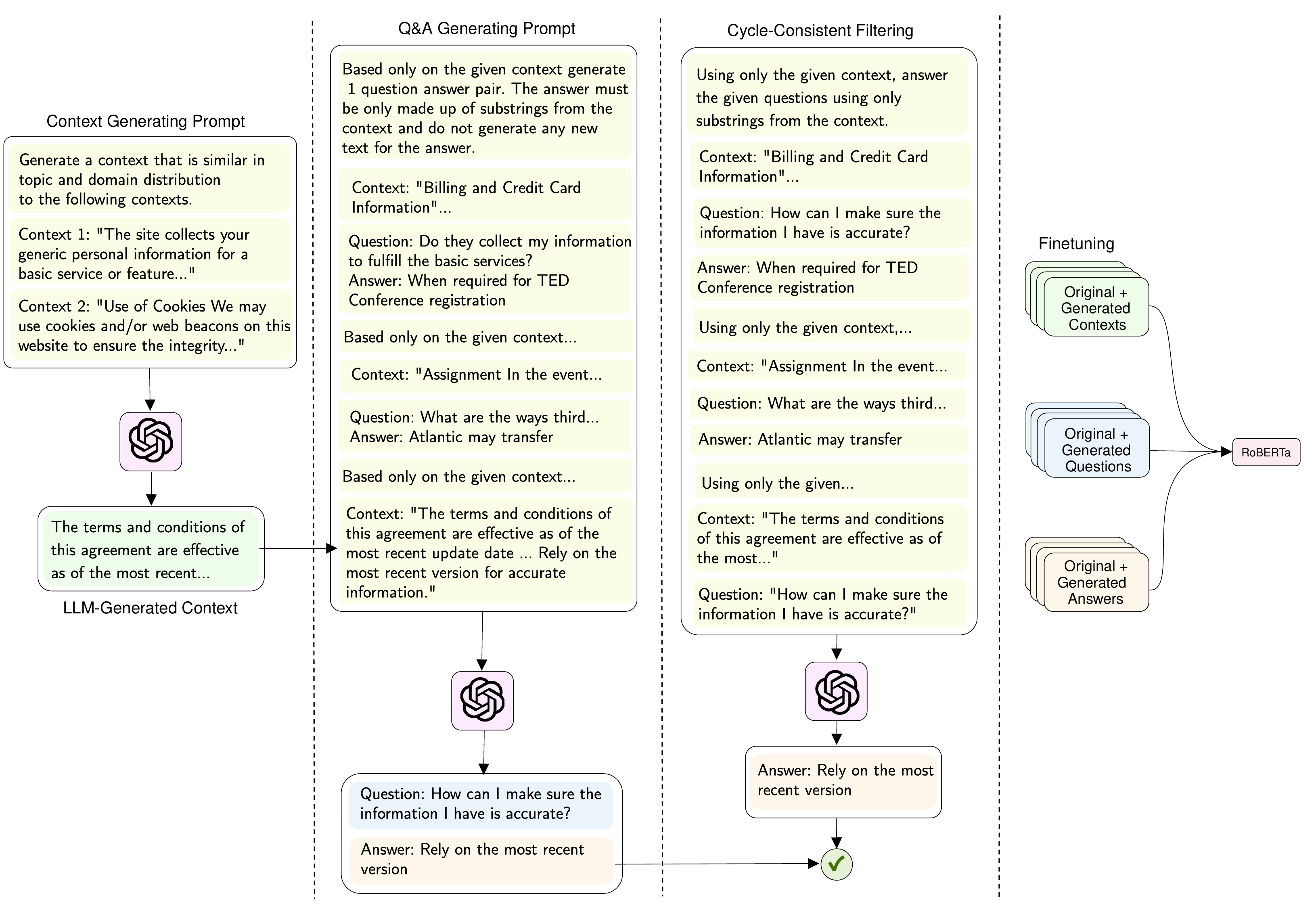}
 \caption{Overview of our methodology using PolicyQA as an example with 2-shot prompts.}
 \label{fig:model}
\end{figure*}

\section{Synthetic Data Generation}

We generate synthetic examples for each dataset using the in-context learning capabilities of the GPT-4 model. The data generation process consists of two stages:

\subsection{Context Generation}

In the first stage, we provide GPT-4 with either 1 example (one-shot) or 2 examples (two-shot) of contexts from the original training set of each dataset. These few-shot examples prime GPT-4 on the style and topics present in the contexts. Providing just one or two examples allows GPT-4 to adapt from demonstrations due to the robust few-shot learning capabilities of LLMs \cite{reif-etal-2022-recipgpt_in_context_1, frohberg-binder-2022-crass_in_context_2, wei2022chain_cot_prompting_in_context_3}. We then generate new synthetic paragraph-length contexts by providing a prompt and allowing GPT-4 to complete the paragraph based on the few-shot priming.

\subsection{QA Generation}

The second stage generates synthetic question-answer pairs conditioned on the synthetic contexts. We again prime GPT-4 with either 1 example (one-shot) or 2 examples (two-shot) of QA pairs from the original dataset. The few-shot priming allows GPT-4 to learn the QA pattern quickly. We then provide the synthetic context from the first stage along with a prompt for GPT-4 to generate a relevant question and answer pair mimicking the style of the examples.

This two-stage process allows us to leverage the few-shot learning and text generation capabilities of GPT-4 to produce synthetic datasets that mimic the style and semantics of the original data. We generate varying amounts of synthetic data, from 1x to 10x the size of the original training sets, to study the impact on downstream task performance.

\subsubsection{Round Trip Filtration}

To further improve the quality of the synthetic QA pairs, we implement a round trip filtration technique. After generating a synthetic question and answer using GPT-4, we provide the question back to the model without the answer. We allow GPT-4 to attempt answering the question again based on the context. If the model's newly generated answer matches the original synthetic answer, we retain this QA pair, as it indicates a high quality question with a consistent answer. If the answers do not match, we discard the synthetic QA pair under the assumption that the question is flawed in some way.

This round trip filtration process provides a mechanism for GPT-4 to self-filter its own generated content. By only keeping QA pairs that exhibit consistency when answered twice, we obtain higher quality synthetic data for downstream training. The filtration process improves precision at the potential expense of some recall.

\subsection{Experiments}

We train an extractive reading comprehension model, using the RoBERTA-Base model across all our experiments. We use a learning rate of $3e-5$, a batch size of $16$ and run our experiments for $5$ epochs each. We use the implementation provided by HuggingFace, and run our models on a stand-alone Nvidia V100 GPU. For all our experiments, we measure F1 and Exact Match scores.

As a baseline for question-answer generation we use a T5 based question generation model that is trained on the SQUAD dataset, which takes a paragraph has an input and returns a question-answer pair. We use the open source \footnote{\href{https://github.com/patil-suraj/question\_generation}{https://github.com/patil-suraj/question\_generation}} implementation for this model.

\begin{table}[]
    \centering
          \resizebox{0.8\columnwidth}{!}{    
    \begin{tabular}{l|c|c}
        \toprule
        \multicolumn{3}{c}{\textbf{CovidQA}} \\
        \midrule
        \acb{\textbf{Setup}} & \textbf{Exact Match} & \textbf{F1 Score} \\
        \midrule
        Original Trainset & 25.81 & 50.91 \\
        Baseline & 19.71 & 44.18 \\
        One Shot & 30.82 & 57.87 \\
        Two Shot & 31.18 & 55.64 \\
        One Shot (CC) & \textbf{31.90} & \textbf{58.66} \\
        Two Shot (CC) & 30.82 & 53.40 \\
    \end{tabular}
    }
    \vspace{0.2em}

              \resizebox{0.8\columnwidth}{!}{
    \begin{tabular}{l|c|c}
        \toprule
        \multicolumn{3}{c}{\textbf{PolicyQA}} \\
        \midrule
        \acb{\textbf{Setup}} & \textbf{Exact Match} & \textbf{F1 Score} \\
        \midrule
        Original Trainset & 30.56 & 58.15 \\
        Baseline & 30.08 & 57.65 \\
        One Shot & \textbf{32.18} & \textbf{59.61} \\
        Two Shot & 30.97 & 59.12 \\
        One Shot (CC) & 30.76 & 58.71 \\
        Two Shot (CC) & 30.47 & 58.46 \\
    \end{tabular}
    }
    \vspace{0.2em}

              \resizebox{0.8\columnwidth}{!}{
    \begin{tabular}{l|c|c}
        \toprule
        \multicolumn{3}{c}{\textbf{TechQA}} \\
        \midrule
        \acb{\textbf{Setup}} & \textbf{Exact Match} & \textbf{F1 Score} \\
        \midrule
        Original Trainset & 11.11 & 39.45 \\
        Baseline & \textbf{44.44} & \textbf{59.92} \\
        One Shot & 22.22 & 36.91 \\
        Two Shot & 11.11 & 36.50 \\
        One Shot (CC) & 22.22 & 41.76 \\
        Two Shot (CC) & 22.22 & 44.73 \\
        \bottomrule
    \end{tabular}
    }
    
    \caption{Experimental Results for MRC Across Various Datasets and Settings.}
    \label{Tab1}
\end{table}

\section{Results}

Table \ref{Tab1} highlights results across the three datasets. For the CovidQA dataset, we observed steady improvements in question answering performance as we augmented the original training set with increasing amounts of synthetic data generated by GPT-4. Using just the original training examples, our model achieved baseline exact match (EM) and F1 scores on the validation set. Adding one-shot synthetic examples improved both the EM and F1 metrics over the baseline. We observed further gains when using two-shot synthetic data, achieving higher EM and F1 compared to one-shot.

The best validation results on CovidQA were obtained by using the one-shot synthetic dataset combined with the round trip filtration process. This achieved the highest EM and F1 scores, significantly improving over the original training distribution. We hypothesize that the round trip filtration allows for higher precision synthetic data, while the one-shot generation provides greater diversity compared to two-shot. The balance of quality and variety in this one-shot filtered dataset appears optimal for augmenting the limited original examples in the CovidQA training set.

In summary, for the CovidQA task we find that synthetic data augmentation uniformly improves performance as more examples are added. The best results come from combining one-shot generation with round trip filtration, which improves exact match and F1 score over the baseline set using just the original dataset. 

With over 12,000 examples, PolicyQA was the largest dataset we utilized. For this task, augmenting the original training set with one-shot synthetic data without filtration achieved the best question answering performance. This improved exact match by 1.6 points and F1 score by 1.5 points compared to using just the original examples. The one-shot augmentation outperformed both two-shot and cycle filtered variations.

Overall for PolicyQA, we find that synthetic data augmentation consistently improves upon the baseline set using just the original training examples. The best configuration utilizes \acb{unfiltered} one-shot generation, likely due to the greater diversity of examples compared to two-shot or filtered versions. While the domain of US immigration policies has high complexity, the large size of the PolicyQA dataset reduces the need for precision-enhancing filtration. The additional synthetic examples provide useful variability when training the model.

With only 1,808 examples, TechQA was the smallest dataset in our experiments. The tiny test set of just 9 examples also made evaluation challenging. On this task, augmenting with synthetic data did not lead to clear improvements in question answering accuracy over the original training set. The baseline model trained on just the 1,808 TechQA examples achieved the highest exact match score, with the two-shot cycle filtered, one-shot filtered, and one-shot unfiltered configurations performing second best in terms of EM. For F1, two-shot cycle filtered data obtained the second highest score after the baseline.

The lack of consistent gains from synthetic data augmentation on TechQA can likely be attributed to the very small data size. With fewer than 2,000 training examples, there is insufficient context for the language model to learn effective generalization. The technical support domain also exhibits diversity that may not be captured from only 1-2 conditioning examples. Furthermore, the small test set provides high variance in evaluation.

\section{Opportunities}

Our experiments demonstrate the significant potential of leveraging large language models (LLMs) like GPT-3 for synthetic data generation. In the CovidQA and PolicyQA domains where a moderate amount of training data was available, augmenting with LLM-produced synthetic examples consistently improved performance over the baseline trained on just the original dataset. This confirms the few-shot generalization abilities of modern LLMs in producing varied, high-quality synthetic data when primed with only a handful of real examples. Indeed, the one-shot synthetic data augmented models achieved the best results on both CovidQA and PolicyQA, surpassing two-shot and other configurations.

The natural language generation capabilities of LLMs afford great opportunity to increase the diversity and size of limited training sets for downstream tasks. By prompting the models to produce synthetic examples mimicking the patterns in the data, we can expand datasets to be orders of magnitude larger with plausible, human-like samples. This data augmentation approach can be applied to many NLP tasks suffering from small training sizes like reading comprehension, summarization, translation, and more. High-quality synthetic data translates into better task performance without the expense of human labeling efforts.

Critical research directions include developing more advanced filtering techniques to distill only the most useful synthetic samples, as well as integrating external knowledge sources to improve few-shot priming. But the overarching opportunity is clear -- properly harnessed, LLMs have enormous potential to ameliorate the limited data problem through strategic synthetic generation.

\section{Challenges}

However, our experiments on the extremely small TechQA dataset also reveal current limitations in using LLMs for robust synthetic data generation. When provided with only around 1,000 original training examples, the LLM-augmented models performed no better than baseline. The models failed to learn adequate representations from such scarce data for producing useful synthetic examples. This highlights how modern LLMs, despite their progress, still struggle in low-data regimes where broad generalization capabilities are required.

Critical challenges remain in improving LLMs' few-shot learning to make them reliable across diverse domains. Environments with limited data require synthesizing examples from broader conceptual knowledge, not just mimicking surface patterns. Integrating external knowledge into LLMs is an active area of research, but effectively utilizing such knowledge in few-shot scenarios remains difficult. There are also challenges in filtering large volumes of synthetic data to maximize diversity while maintaining precision and quality.

In summary, while LLMs offer promise for alleviating limited training data, substantial challenges persist. Robustness to low-data regimes, integration of world knowledge, and advanced content filtering mechanisms are needed to make synthetic data generation truly effective for any NLP task. This is an exciting and rapidly evolving area of research that will determine whether LLMs can deliver on their potential to mitigate limited datasets through strategic synthetic example construction.

\bibliography{anthology,custom}
\bibliographystyle{acl_natbib}

\appendix



\end{document}